\documentclass{article}
\usepackage{spconf}
\usepackage{amsmath,graphicx}
\usepackage{amssymb}
\usepackage[dvipsnames]{xcolor}
\usepackage{textcomp}
\usepackage{tabularx}
\usepackage{algpseudocode}
\usepackage{algorithm}
\usepackage{amsfonts}
\usepackage[colorinlistoftodos]{todonotes}
\usepackage{booktabs}
\usepackage[export]{adjustbox}
\usepackage{array}
\usepackage{float}
\usepackage{multirow}
\usepackage[none]{hyphenat}
\usepackage{tikz}
\usepackage{subcaption}
\usepackage{lipsum}
\usepackage{siunitx}
\usetikzlibrary{bayesnet}
\usetikzlibrary{arrows}
\usetikzlibrary{backgrounds}
\usetikzlibrary{positioning}

\def\x{{\mathbf x}}
\def\N{{\mathcal{N}}}

\def\D{{\mathcal{D}}}

\def\bfphi{{\boldsymbol{\phi}}}
\def\veta{{\boldsymbol{\eta}}}

\title{Fusion of Gaussian Processes Predictions with Monte Carlo Sampling}
\name{Marzieh Ajirak, Daniel Waxman, Fernando Llorente,   and Petar M. Djuri\'c
\thanks{This work was supported by the National Science Foundation under Award 2212506.}}
\address{Department of Electrical and Computer Engineering \\
Stony Brook University
}
\ninept

\begin{document}
\copyrightnotice{\copyright\ 2023 IEEE}
\maketitle

\begin{abstract}
  In science and engineering, we often work with models designed for accurate prediction of variables of interest. Recognizing that these models are approximations of reality, it becomes desirable to apply multiple models to the same data and integrate their outcomes. In this paper, we operate within the Bayesian paradigm, relying on Gaussian processes as our models. These models generate predictive probability density functions (pdfs), and the objective is to integrate them systematically, employing both linear and log-linear pooling. We introduce novel approaches for log-linear pooling, determining input-dependent weights for the predictive pdfs of the Gaussian processes. The aggregation of the pdfs is realized through Monte Carlo sampling, drawing samples of weights from their posterior. The performance of these methods, as well as those based on linear pooling, is demonstrated using a synthetic dataset.
\end{abstract}

\section{Introduction}

In science and engineering, dealing with models that aim to accurately predict a variable of interest is a common task. Since these models often are only approximations of reality, it is prudent to apply multiple models to the same data and somehow integrate their predictions.  
Within the Bayesian framework, results based on models are typically presented as predictive probability distributions --- in the case of continuous variables, they are commonly in the form of predictive probability density functions (pdfs). In such settings, it is important to combine the produced predictive pdfs \cite{koliander2022fusion}. In this paper, we focus on models defined by Gaussian processes (GPs) and propose an approach to fuse the predictive pdfs of an {\em ensemble} of different GPs. This involves aggregating Gaussians obtained through various instances of linear and log-linear pooling of the Gaussian pdfs. The different instances correspond to different weights assigned to the GPs, where the weights are drawn by Monte Carlo sampling from the posterior of the weights.

GPs are probabilistic non-parametric models that find the function that probabilistically maps an input variable $\mathbf{x}$ to an output variable $y$. The density of the output $y$ is Gaussian with parameters that depend on $\mathbf{x}$. This model is inaccurate when the output is not Gaussian.
The effectiveness of model ensembles in enhancing predictive performance has long been recognized \cite{dietterich2000ensemble}. However, understanding the conditions under which an ensemble of GPs is likely to yield reliable estimates is not straightforward. Bayesian Model Averaging (BMA) \cite{bernardo2009bayesian} operates under the assumption that the true model lies within the hypothesis class of the prior. However, its suboptimality outside this case is shown empirically in \cite{domingos2000bayesian}, with an extreme (synthetic) example provided in \cite{minka2000bayesian}.

In contrast, ensembles adopt a different strategy by focusing on model combinations. By merging multiple models, ensembles create a more robust and powerful model. Ensembles demonstrate particular efficacy when the actual model lies beyond the hypothesis class. In such instances, the collective strength derived from the diverse models contributes to an overall improvement in predictive performance.

One of the most popular ensembling methods for point estimators is \textit{stacking} \cite{wolpert1992stacked}, which employs $K$ point estimators denoted as $f_k \colon \mathbf{x} \mapsto y_k$. The stacking process involves training an additional point estimator whose inputs are the outputs of each $f_k$, represented as $\hat{y} = g(\hat{y}_1, \dots, \hat{y}_K)$. Stacking is quite general, and in particular, $g$ can be determined in many ways, such as using a linear model with weights $w_k$, or more sophisticated functions like neural networks. 

% Adapting the stacking approach to a Bayesian formalism has been attempted in several different ways. In \textit{Bayesian classifier combination} \cite{kim2012bayesian}, a Bayesian model is proposed to model the errors of classifiers, deriving a Bayesian posterior over a fused estimate.
\emph{Bayesian stacking} \cite{yao2018using} adapts the stacking approach to probabilistic estimators by replacing point estimates $y_k$ with predictive pdfs $p_k(y_* \,\lvert\, {\bf x}_*)$ and $\hat{y}$ with a mixture $\sum w_k p_k(y_* \,\lvert\, {\bf x}_*)$. One step further is Bayesian hierarchical stacking \cite{yao2022bayesian}, which incorporates a Bayesian prior on input-dependent weights, making the final mixture $\sum \hat{w}_k({\bf x}_*) p_k(y_* \,\lvert\, {\bf x}_*)$. 

Distinct from stacking is the \emph{mixture of experts (MoE)} method \cite{jacobs1991adaptive}, which forms a mixture by jointly learning several experts, and a gating network that determines the weights assigned to the experts. This method is generally more expensive and difficult to train than stacking, but it allows for more expressive predictive pdfs.
% Moreover, \textit{hierarchical mixtures of experts (HMEs)} \cite{jordan1994hierarchical}  have been used in numerous fusion applications. These models consider a conditional mixture models, representing target variable distributions through mixtures of expert distributions. The expert distributions and mixing coefficients are conditioned on input variables, offering a flexible framework for capturing complex relationships in diverse domains.

In Section~\ref{sec_2}, we present a brief background on GPs and Monte Carlo sampling, thus providing both the setting and methodological tools for this paper. In Section~\ref{sec_3}, we describe several existing strategies for the fusion of GP predictions, namely Bayesian hierarchical stacking (BHS) and the mixture of GP experts (MoGPE). We then introduce novel variants of BHS and MoGPE by replacing their linear pooling with log-linear pooling, resulting in the Product BHS (P-BHS) and the product of GP experts (PoGPE). Finally, we show how all four models can be estimated with Monte Carlo sampling and random Fourier feature-based GPs (denoted as RFF-GPs), and we provide numerical comparisons between methods.

\section{Background}\label{sec_2}
We briefly review the basics of GP regression, random Fourier feature approximation of kernel functions, and Bayesian inference using Monte Carlo sampling.

\subsection{Gaussian Processes}
A GP is described by a stochastic process, where any finite subset of random variables jointly conforms to a Gaussian distribution \cite{williams2006gaussian}. A GP is fully specified by its mean $\mu(\mathbf{x})$ and covariance function $\kappa\left(\mathbf{x}, \mathbf{x}^{\prime}\right)$ with hyperparameters $\boldsymbol{\theta}$.
In training a GP, we seek hyperparameters $\boldsymbol{\theta}$ that maximize the log-marginal likelihood of the hyperparameters defined by 
\begin{align}
\log p(\mathbf{y} \,\lvert\, \mathbf{X}, \boldsymbol{\theta}) = -\frac{1}{2}\left(\mathbf{y}^T \mathbf{K}^{-1} \mathbf{y} + \log \lvert\mathbf{K}\rvert\right),
\end{align}
where ${\bf K} \in \mathbb{R}^{N \times N}$ is a covariance matrix with elements $\left[\mathbf{K}\right]_{ij} = \kappa(\mathbf{x}_i, \mathbf{x}_j)$. Alternatively, we can place a prior $p(\boldsymbol{\theta})$ on the kernel hyperparameters and proceed under the Bayesian paradigm.

For a given set of hyperparameters $\boldsymbol{\theta}$, a training set $\mathbf{X}, \mathbf{y}$, and a test input $\mathbf{x}_* \in \mathbb{R}^{d_x}$, the GP posterior predictive pdf of the corresponding function value $f_*=f\left(\mathbf{x}_*\right)$ is Gaussian with mean and variance given by:
\begin{align}
\mathbb{E}\left[f_*\right] & = m\left(\mathbf{x}_*\right) = \mathbf{k}_*^T \mathbf{K}^{-1} \mathbf{y}, \label{eq_GP_postMean} \\
\operatorname{var}\left[f_*\right] & = \sigma^2\left(\mathbf{x}_*\right) = k_{* *} - \mathbf{k}_*^T \mathbf{K}^{-1} \mathbf{k}_*, \label{eq_GP_postVar}
\end{align}
respectively, where $\mathbf{k}_* =\kappa\left(\mathbf{X}, \mathbf{x}_*\right)$ and $k_{* *} = \kappa\left(\mathbf{x}_*, \mathbf{x}_*\right)$.

\subsection{Random Fourier Feature-based Gaussian Processes}
By the very nature of being non-parametric, the number of parameters in a GP grows with the data. This can easily be seen by reparameterizing the GP prior for a fixed training set $\mathbf{X}$,
\begin{align}
    f(\mathbf{X}) &= \mathbf{L} \boldsymbol{u}, \\
    \boldsymbol{u} &\sim \mathcal{N}(\mathbf{0}, \mathbb{I}_{N}),
\end{align}
where $\mathbf{L} = \mathrm{chol}(\mathbf{K})$ is the Cholesky decomposition of the kernel matrix and $\mathbb{I}_N$ is the $N\times N$ identity matrix. 
Moreover, computation of the posterior predictive mean and variance requires the inversion of an $N \times N$ matrix (cf. Eqs. \eqref{eq_GP_postMean} and \eqref{eq_GP_postVar}). The high dimensionality makes sampling difficult for large datasets, and the matrix inversion makes prediction expensive. 

If the power spectral density $S(\omega)$ of the GP kernel exists, one way to alleviate these issues is to take a Monte Carlo approximation of the kernel by drawing frequencies from the normalized power spectral density of the desired kernel and using them with the input data to form features. We do this by  sampling $M$ sets of \emph{random frequencies} from the power spectral density to approximate the integral
\begin{equation}
    \kappa(\x, \x') = \frac{1}{2\pi}\int_{\boldsymbol{\omega}} e^{i\boldsymbol{\omega}^\top ( \x - \x' )} S(\boldsymbol{\omega}) \, d\boldsymbol{\omega},
\end{equation}
where $\boldsymbol{\omega}$ is a vector of frequencies of the same size as  ${\bf x}$. 
The resulting model is a linear model with $2M$ parameters; therefore, if $2M < N$, we can expect computational savings. For general kernel machines, this approximation (known as the \emph{random Fourier feature (RFF) approximation}) was introduced by \cite{rahimi2007random} and was adapted to GP regression in \cite{lazaro2010sparse}.

If $\boldsymbol{\omega}_1, \dots, \boldsymbol{\omega}_M$ are the vectors of frequencies sampled from $S(\boldsymbol{\omega})$, we create a feature vector according to %the kernel is approximated as $\mathbf{K} \approx \phi(\mathbf{X})^\top \phi(\mathbf{X})$, where $\phi(\x)$ is 
\begin{equation}
    \bfphi(\x) = [ \cos(\boldsymbol{\omega}_1^\top\x) \, \sin(\boldsymbol{\omega}_1^\top\x) \, \cdots \, \cos(\boldsymbol{\omega}_M^\top\x) \, \sin(\boldsymbol{\omega}_M^\top\x)]^\top.
\end{equation}
Then using this reparameterization, the resulting model is a RFF-GP given by 
\begin{align}
    f(\x) &= \bfphi(\x)^\top \boldsymbol{\psi}, \\
    \boldsymbol{\psi} &\sim \mathcal{N}(\mathbf{0}, \sigma^2_{\psi}\mathbb{I}_{2M}),
\end{align}
where we assume that the vector of linear parameters $\boldsymbol{\psi}$ has a zero mean Gaussian prior with a covariance matrix $\sigma^2_{\psi}\mathbb{I}_{2M}$.  
%We refer to the GPs using the random Fourier feature approximation as \emph{r-GPs}.

% \begin{table}
%     \centering
%     \begin{tabular}{ccl}
%         \toprule
%         $\mathcal{GP}$ & -- & Gaussian process\\
%  $\kappa$& --&Covariance function of the Gaussian process\\
%          $K$ & --  & Number of GP Experts\\
%          &  & \\
%          \bottomrule
%     \end{tabular}
%     \caption{Caption}
% \end{table}

\subsection{Bayesian Inference via Monte Carlo Algorithms}

In Bayesian inference, the goal is to infer the parameters of a statistical model for the observations by computing the posterior distribution.
Let $\veta$ be a parameter vector, and let   $p({\bf y}\,\lvert\,\veta,{\bf X})$ and $p(\veta)$ be the likelihood and the prior of $\veta$, respectively. 
As we will see below, in this work, $\veta$ contains all the parameters of the RFF-GPs. % , $\veta = [\boldsymbol{\psi}_1,\boldsymbol{\psi}_2,\dots]$, and other hyperparameters such as $\sigma_{\psi_k}$.
The posterior 
$p(\veta\,\lvert\,{\bf y},{\bf X}) \propto p({\bf y}\,\lvert\,\veta,{\bf X})p(\veta)$ is usually intractable, and we need to resort to approximate inference methods, such as Markov chain Monte Carlo (MCMC) which computes an approximation based on samples \cite{Robert04}. 
In our work, we use Hamiltonian Monte Carlo (HMC) and its adaptively tuned variant ``No-U-Turn Sampler'' (NUTS) \cite{duane1987hybrid}, which are state-of-the-art gradient-based MCMC algorithms.
Given $N_{s}$ samples $\{\veta^{(i)}\}_{i=1}^{N_{s}}$ from $p(\veta\,\lvert\,{\bf y},{\bf X})$, we can approximate the posterior predictive pdf of a new observation $y_*$ as follows:
\begin{align}
p(y_*\,\lvert\,\x_*,{\bf y}, {\bf X}) \approx \frac{1}{N_{s}}\sum_{i=1}^{N_{s}} p(y_*\,\lvert\,\x_*,\veta^{(i)}),\quad \veta^{(i)} \sim p(\veta\,\lvert\,{\bf y},{\bf X}).
\end{align}

\section{Fusion of Gaussian processes predictions}
\label{sec_3}

\begin{figure}
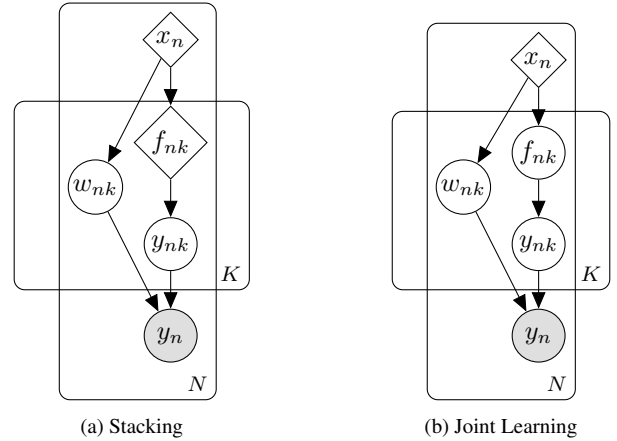

\centering
\hfill

\begin{subfigure}[b]{0.44\columnwidth}
    \centering
      \tikz{
\node[obs] (y) {$y_n$};%
\node[latent,above=0.5 of y] (yk) {$y_{nk}$}; %
\node[det,above=0.5 of yk] (theta) {$f_{nk}$}; %
\node[latent,above=1.25 of y, xshift=-1 cm] (wk) {$w_{nk}$}; %
\node[det,above=0.5 of theta] (x) {$x_n$}; %
\node[right=0.35 of yk] (ykdummy) {};
\node[left=1.25 of theta] (thetadummy) {};
\node[above=0.1 of thetadummy] (thetadummy2) {};

\edge {yk} {y};
\edge{theta}{yk};
\edge{wk}{y};
\edge{x}{theta};
%\path (x) edge[->, >={triangle 45}, bend left] (yk);
\edge {x} {wk};

\plate{plate1} {(theta)(y)(yk)(wk)(x)} {$N$};
\plate{plate2}{(thetadummy2)(wk)(ykdummy)}{$K$};
  }
  \caption{Stacking}
   \end{subfigure}
\hfill
\begin{subfigure}[b]{0.4\columnwidth}
    \centering
      \tikz{
\node[obs] (y) {$y_n$};%
\node[latent,above=0.5 of y] (yk) {$y_{nk}$}; %
\node[latent,above=0.5 of yk] (theta) {$f_{nk}$}; %
\node[latent,above=1.25 of y, xshift=-1 cm] (wk) {$w_{nk}$}; %
\node[det,above=0.5 of theta] (x) {$x_n$}; %
\node[right=0.25 of yk] (ykdummy) {};
\node[left=1.25 of theta] (thetadummy) {};
\node[above=0.1 of thetadummy] (thetadummy2) {};

\edge {yk} {y};
\edge{theta}{yk};
\edge{wk}{y};
\edge{x}{theta};
%\path (x) edge[->, >={triangle 45}, bend left] (yk);
\edge {x} {wk};

\plate{plate1} {(theta)(y)(yk)(wk)(x)} {$N$};
\plate{plate2}{(thetadummy2)(wk)(ykdummy)}{$K$};
  }
  \caption{Joint Learning}
   \end{subfigure}
   \hfill
   \caption{Bayesian plate diagrams of joint learning and stacking for Gaussian process fusion. The only difference is whether the stochastic function $f_k$ is treated as an input to be conditioned on (as in \protect{(a)}), or a random variable (as in \protect{(b)}).
   }\label{fig:bayes_nets}
\end{figure}

Let us consider the case where we have $K$ different GPs, each with a predictive pdf $\mathcal{N}\big(y \,\lvert\,\mu_k(\x), \sigma^2_k(\x)\big)$ of $y$ conditioned on $\x$. We aim to combine these predictive pdfs, $p_k(y \,\lvert\,\x)$ for $k=1,\dots, K$ to a single, fused pdf $p_{\text{f}}(y \,\lvert\,\x)$. In this section, we review the concepts of \emph{Bayesian hierarchical stacking} (Section~\ref{sec:bhs}) and the \emph{mixture of GP experts} (Section~\ref{sec:mogpe}), which both use input-dependent weights $\mathbf{w}(\x) \in \mathbb{S}_K$ and a linear pooling function to result in a fused pdf $p_{\text{f}}(y \,\lvert\,\x) = \sum w_k(\x) p_k(y \,\lvert\,\x)$. We then present a novel version of both stacking and mixture of GP experts by using a log-linear pooling function (Section~\ref{sec:our_method}).

\subsection{Bayesian Hierarchical Stacking} \label{sec:bhs}
In \emph{Bayesian hierarchical stacking (BHS)} \cite{yao2022bayesian}, one fuses $K$ different predictive pdfs from pre-trained algorithms. We assume that these predictive pdfs come from $K$ GPs, and they are  $p_k(y \,\lvert\,\x) = \mathcal{N}\big(y \,\lvert\,\mu_k(\x), \sigma^2_k(\x)\big),$ all obtained by using Eqs. \eqref{eq_GP_postMean} and \eqref{eq_GP_postVar}. BHS creates fused predictive pdfs using a linear pooling function, i.e.,
\begin{equation}
    p_{\text{lin}}\big(y \,\lvert\,\x, \mathbf{w}(\x)\big) = \sum_{k=1}^K w_k(\x) \mathcal{N}\big(y \,\lvert\,\mu_k(\x), \sigma^2_k(\x)\big),
\end{equation}
where $\mathbf{w}(\x)$ belongs to the simplex $\mathbb{S}_K$, i.e., $\sum_{k=1}^K w_k(\x) = 1$ for all $\x$. 

BHS lays out a general framework to obtain these weights, but for our work, we learn the input-dependent weights with an additional GP trained on a {\em separate} stacking dataset $\mathcal{D}_s$. The full model becomes \cite[Section~5.2]{yao2022bayesian},
\begin{align}
\widetilde{w}_k(\mathbf{x}) &\sim \mathcal{GP}(0,\kappa_k(\mathbf{x},\mathbf{x}')), \enskip \widetilde{w}_K(\x) = 0, \;\; k=1,\dots,K-1,  \\
     {\bf w}(\mathbf{x}) &= \text{softmax}(\widetilde{{\bf w}}(\mathbf{x})),  \\
    y \,\lvert\, {\bf w}(\mathbf{x}), \mathbf{x} &\sim \sum_{k=1}^K w_k(\mathbf{x})\N\big(y \,\lvert\,\mu_k(\x), \sigma_k^2(\x)\big).
    \label{eq_bhs_lik}
\end{align}

We can perform inference with this model by sampling from the posterior $p({\bf w}(\cdot) \,\lvert\,\D_s)$ via HMC.
The final predictive pdf for an input $\x_*$ is
\begin{align}
    p_{\text{BHS}}(y_*\,\lvert\,\x_*) = \sum_{k=1}^K \bar{w}_k(\x_*) \N(y\,\lvert\,\mu_k(\x_*), \sigma_k^2(\x_*)),
\end{align}
where $\bar{w}_k(\x_*)$ denotes the MMSE estimate/posterior mean of $w_k(\cdot)$ at $\x_*$. Because of linearity of the integral and the independence of $w_k$ from $y \,\lvert\, \mu_k, \sigma_k$, the use of the posterior mean $\bar{w}_k$ provides the same estimator as the full approximation
\begin{equation}
    p_{\text{BHS}}(y_*\,\lvert\,\x_*) = \frac{1}{N} \sum_{n=1}^N \sum_{k=1}^K w_{k,n}(\x_*) \N(y\,\lvert\,\mu_k(\x_*), \sigma_k^2(\x_*)).
\end{equation}

\subsection{Mixture of GP Experts} \label{sec:mogpe}
An important feature of the BHS strategy is that the $K$ GPs are fixed during the learning of the weights. In other words, the functions $\mu_k(\x)$ and $\sigma^2(\x)$ are computed using an initial training dataset $\mathcal{D}_0$ and kept fixed from there on. To create a more flexible model, we can instead use a single dataset $\mathcal{D}_{\text{train}}$ to perform joint training of the $K$ GP experts and the weights. This joint learning falls into the mixture of experts framework \cite{jacobs1991adaptive}, and was first proposed for GPs in \cite{tresp2000mixtures}. 

In order to have flexible experts, \cite{tresp2000mixtures} uses GP experts with input-dependent noise; this can be performed by placing GP priors on both the mean $\mu_k(\x)$ and log-scale $\log\sigma_k(\x)$ \cite{goldberg1997regression}. Again using the linear pooling function, we arrive at the \emph{mixture of GP experts} (MoGPE) defined by
\begin{align}
      \mu_k(\x) &\sim \mathcal{GP}(0,\kappa_k^{\mu}),\ k=1,\dots,K,  \\
      \log\sigma_k(\x) &\sim \mathcal{GP}(0,\kappa_k^{\sigma}),\ k=1,\dots,K,  \\
    \widetilde{w}_k(\mathbf{x}) &\sim \mathcal{GP}(0,\kappa^w_k), \enskip k=1,\dots,K-1;  \widetilde{w}_K(\x) = 0,  \\
     {\bf w}(\mathbf{x}) &= \text{softmax}(\widetilde{{\bf w}}(\mathbf{x})),  \\
    y  \,\lvert\, {\bf z}, \x  &\sim \sum_{k=1}^K w_k(\mathbf{x})\N(y\,\lvert\,\mu_k(\x), \exp(2\log\sigma_k(\x)) ), \label{eq_GPmoe_lik}
\end{align}
where ${\bf z} = \{w_k(\x),\mu_k(\x),\log\sigma_k(\x)\}_{k=1}^K$.
In order to obtain the final predictive pdf, we marginalize the likelihood in Eq. \eqref{eq_GPmoe_lik} with respect to the posterior of $w_k(\cdot)$, $\mu_k(\cdot)$, and $\log\sigma_k(\cdot)$ for $k=1,\dots, K$.

We can again sample the unknowns of the model with HMC. However, the resulting posterior of the unknowns is much more complex to sample than the BHS posterior --- as a result, we propose using RFF GPs for means, log-scales, and non-normalized weights for scalable inference.

\subsection{Log-Linear Pooling for Stacking and Joint Learning} \label{sec:our_method}
While the BHS and MOGPE methods focused on linear pooling of GP predictions, one can instead use a log-linear pooling rule, where a weighted average of the log-pdfs is taken instead of the pdfs. The log-linear pooling scheme was first analyzed in \cite{genest1984characterization}, with an axiomatic characterization and comparison available in \cite{koliander2022fusion}, and it possesses analytically different properties than linear pooling. As the weighted average of log-pdfs is equivalent to a weighted geometric average, we can express the resulting pdf as a weighted product,
\begin{equation} \label{eq:log_linear}
    p_{\text{log-lin}}(y \,\lvert\,\x) = c\big(\mathbf{w}(\x)\big) \prod_{k=1}^K p_k(y \,\lvert\,\x)^{w_k(\x)},
\end{equation}
where $c\big(\mathbf{w}(x)\big)$ is a normalizing constant, and $\mathbf{w}$ is typically taken to be in the simplex $\mathbb{S}_K$.

When the predictive pdfs $p_k(\cdot)$ are Gaussian, it is well known that the pdf resulting from Eq.~\eqref{eq:log_linear} is proportional to another Gaussian. In the context of fusing GPs, the resulting rule is known as the \emph{generalized product of experts (gPoE)} \cite{cao2014generalized}, and it is defined by
\begin{align}
    p_{\text{gPoE}}(y \,\lvert\,\x) &= \mathcal{N}\big(y \,\lvert\,\mu_{\text{gPoE}}, \sigma^2_{\text{gPoE}}\big), \label{eq:gpoe_density} \\
    \mu_{\text{gPoE}}(\mathbf{x}) &= \sigma^2_{\text{gPoE}} \sum_{k=1}^K w_k(\x) \sigma_k^{-2}(\x) \mu_k(\x),\\
    \sigma^{-2}_{\text{gPoE}}(\x) &= \sum_{k=1}^K w_k(\x) \sigma^{-2}_k(\x).
\end{align}
Several strategies for setting the weights $\mathbf{w}$ have been proposed in the literature \cite{cao2014generalized,deisenroth2015distributed}. Though these existing strategies have chosen weights belonging to the simplex $\mathbb{S}_K$, this is not necessary for the mathematical validity of Eq. \eqref{eq:gpoe_density}. Instead, it has been chosen for the property that the resulting precision $\sigma_{\text{gPoE}}^{-2}(\x)$ is a convex combination of the expert precisions $\sigma^{-2}_k$.
% Several strategies for setting the weights $\mathbf{w}$ have been proposed in the literature \cite{cao2014generalized,deisenroth2015distributed}. Though these existing strategies have chosen weights belonging to the simplex $\mathbb{S}_K$, this is not necessary for the mathematical validity of Eq. \eqref{eq:gpoe_density}. Instead, it has been chosen for the property that the resulting precision $\sigma_{\text{gPoE}}^{-2}(\x)$ is a convex combination of the expert precisions $\sigma^{-2}_k$.

The existing literature surrounding gPoE has not trained a separate statistical model for input-dependent weights $\mathbf{w}(\cdot)$, nor has it considered joint learning in such a model. To this end, we propose \emph{product-BHS (P-BHS)}, which performs Bayesian hierarchical stacking with a log-linear pooling rule, and the \emph{product of GP experts (PoGPE)}, which performs joint learning.  

Concretely, we propose the following hierarchical model for the input-dependent weights $\mathbf{w}(\cdot)$:
\begin{align}
    \log w_k(\mathbf{x}) &\sim \mathcal{GP}(-\log(K),\kappa_k(\mathbf{x},\mathbf{x}')),\enskip k=1,\dots,K, \\
    y \, \,\lvert\,\, {\bf w}(\mathbf{x}), \mathbf{x} &\sim \mathcal{N}(y\,\lvert\, \mu_\text{gPoE}(\mathbf{x};{\bf w}(\mathbf{x})),\sigma^2_\text{gPoE}(\mathbf{x};{\bf w}(\mathbf{x})) ), \label{eq_our_model_lik}
\end{align}
where we have made explicit the dependence of $\mu_\text{gPoE}$ and $\sigma^2_\text{gPoE}$ on $\mathbf{w}(\x)$. Note that we model the unconstrained weights $\log w_k(\x) \in \mathbb{R}$, meaning we do not restrict weights to $\mathbb{S}_K$ but instead $\mathbb{R}^+$. This has the effect of allowing a heteroscedastic prior predictive of our model, with the use of stationary GP experts. Moreover, the GP prior mean of $-\log K$ implies a prior mean of $1 / K$ on $w_k$. Therefore, we avoid issues of being overconfident as $\x$ falls outside the typical set of our data.

% \begin{table*}[t]
%     \centering
%         \caption{
%     \textcolor{red}{results with $M=30$ spectral frequencies and averaged over 5 different training-test splits}
%     }
%     \begin{tabular}{cccccc}
%     \toprule
%          $K$& 1&2&3& 4 &5\\
%          \,\lvert\,rule
%          BHS& &&&&\\
%          MOGPE& &&&&\\
%          POGPE& &&&&\\
%     PBHS& & & & &\\
%     CDE& &&&&\\
%  \bottomrule
%  \end{tabular}
% \end{table*}

The final model can be written as 
\begin{align}
      \mu_k(\x) &\sim \mathcal{GP}(0,\kappa_k^{\mu}),\ k=1,\dots,K,  \\
      \log\sigma_k(\x) &\sim  
      \mathcal{GP}(0,\kappa_k^\sigma),\\
    \log w_k(\mathbf{x}) &\sim \mathcal{GP}(-\log(K),\kappa^w_k), \enskip k=1,\dots,K, \\
    y \, \,\lvert\,\, {\bf z}, \x &\sim 
    \N(y\,\lvert\, \mu_\text{gPoE}(\mathbf{x};{\bf w}(\x)),\sigma^2_\text{gPoE}(\x;{\bf w}(\x)) ),
\end{align}
where ${\bf z} = \{\mu_k(\x),\log\sigma_k(\x) \log w_k(\x)\}_{k=1}^K$. The variables 
$\mu_k(\x)$ and $\ \log\sigma_k(\x)$ are given in P-BHS and inferred jointly in POGPE. The predictive pdf and its Monte Carlo approximation can then be written as
\begin{align}
    p(y_*\,\lvert\,\mathbf{x}_*)
    % &= 
    % \int \mathcal{N}(y_*\,\lvert\,\mu_\text{gPoE}(\mathbf{x}_*\,\lvert\,{\bf w}(\mathbf{x}_*)),\sigma^2_\text{gPoE}(\mathbf{x}_*\,\lvert\,{\bf w}(\mathbf{x}_*)) ) \nonumber\\
    % &\enskip \enskip \times p({\bf w}(\mathbf{x}_*)\,\lvert\,\mathcal{D}) \, d {\bf w}(\mathbf{x}_*) \nonumber\\
    &\approx\frac{1}{J}\sum_{j=1}^J \mathcal{N}(y_*\,\lvert\,\mu^{(j)}_\text{gPoE},\sigma^{2,(j)}_\text{gPoE}),\label{eq_our_model_pred}
\end{align}
where $\mu^{(j)}_\text{gPoE}$ and $\sigma^{2,(j)}_\text{gPoE}$ are obtained by computing Eq. \eqref{eq:gpoe_density} using the posterior samples of $\{\log w_k(\x_*)\}_{k=1}^K$ in P-BHS or the posterior samples of $\{\mu_k(\x_*), \log\sigma_k(\x_*),\log w_k(\x_*)\}_{k=1}^K$ in PoGPE.

\section{Experiments}
\begin{figure}
    \centering
    \includegraphics[width=\columnwidth]{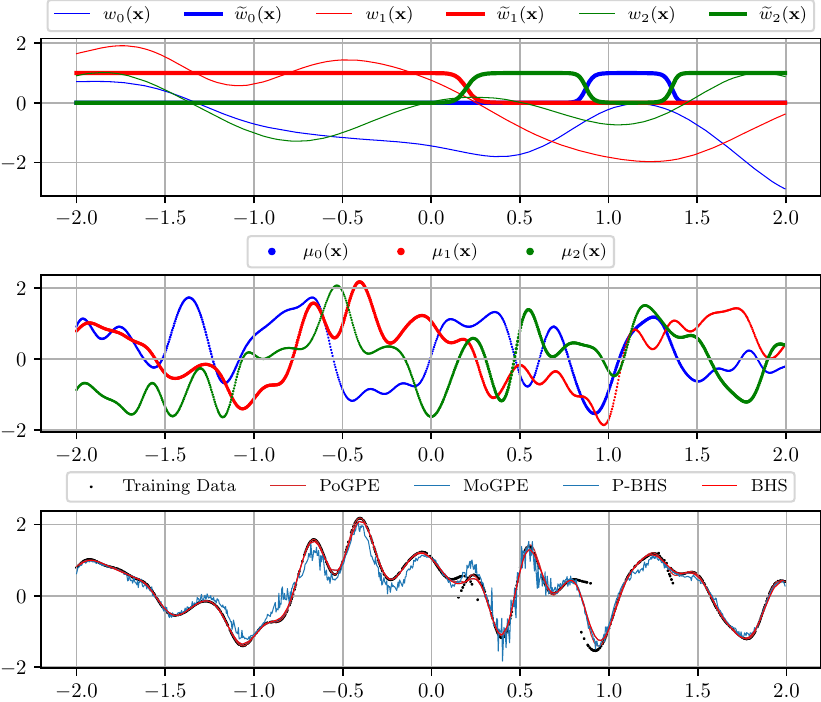}
    \caption{A depiction of the generative model and the point predictions of each method. From top to bottom: the means of each component GP and the respective weight process; sample draws from each GP with linewidth indicating $w_k(x)$; the resulting dataset with the predictions of each method.}
    \label{fig_data}
\end{figure}

In this section, we run a one-dimensional numerical experiment where we test the novel P-BHS and PoGPE. Results are compared to BHS, MoGPE, and a heteroscedastic RFF GP (\emph{het-RFF-GP}). We consider experiments where the number of spectral frequencies $M$ is fixed and the number of experts $K$ is varied (Section~\ref{sec:NLPD_vs_K}), and experiments where $K$ is fixed and $M$ is varied (Section~\ref{sec:NLPD_vs_M}).

We generate data following the generative model in Section~\ref{sec:mogpe} (namely, according to MoGPE). The weights for three functions, $\widetilde{w}_k(\mathbf{x})$, $K=3$ are generated from a GP with zero mean and RBF kernel with lengthscale $\ell=0.5$. Three functions are sampled from another GP with RBF kernel and lengthscale $\ell=0.2$. Given $\x_i$, we first sample the unconstrained weights, apply softmax, and we sample the $i$-th observation from $\N(y\,\lvert\,\mu_k(\x_i), \exp(2\log\sigma_k(\x))$ with probability $w_k(\x_i)$. 
% \textcolor{red}{FA: I didn't sample from the Gaussian, I just gave you the mean but it doesn't matter :/}
Hence, conditional on GP draws, the function of interest is a linear mixture.
The experts, input-dependent weights, and the resulting dataset are shown in Figure~\ref{fig_data}.
We generate \num{1000} data points according to this procedure and consider an 80\%-20\% training-test split, namely, $N_\text{tr}=800$ and $N_\text{tst} = 200$. We compare the methods in terms of their mean negative log-predictive density (NLPD) on test data, averaged over 5 random splits. 
All models were implemented in the probabilistic programming language NumPyro \cite{phan2019composable}, whose NUTS implementation was used for inference. In each case, $500$ samples were drawn from the posterior of each of the four chains, resulting in a total of $J=2000$ samples.

\begin{figure*}[t]
\centering
\hfill
\begin{subfigure}{\columnwidth}
    \centering
    {\includegraphics[width=0.8\columnwidth]{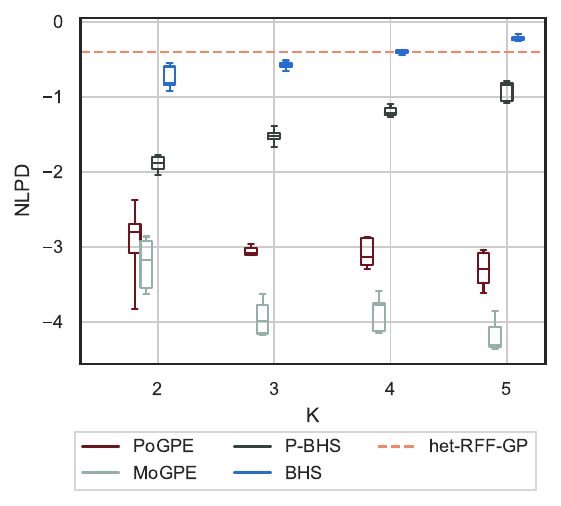}}
    \caption{}
    \label{fig_NLPD_vs_experts}
\end{subfigure}%
\begin{subfigure}{\columnwidth}
\centering
{\includegraphics[width=0.8\columnwidth]{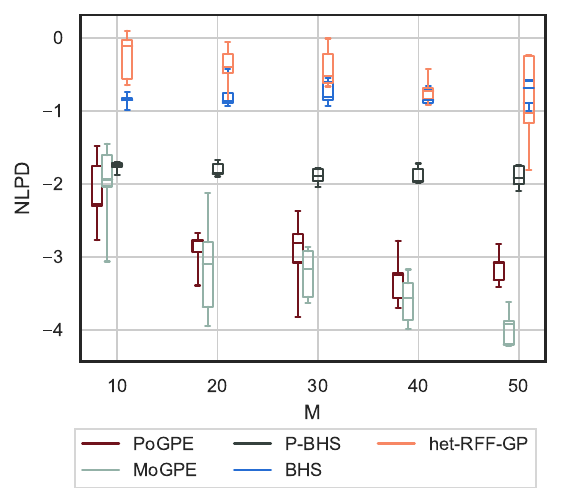}}
    \caption{}
    \label{fig_NLPD_vs_frequencies}
\end{subfigure}
\hfill
\caption{Results from the experiments described in (a) Section~\ref{sec:NLPD_vs_K} and (b) Section~\ref{sec:NLPD_vs_M}. The NLPD \emph{(lower is better)} are averaged across test data; boxes show the median and quartiles of the average across $5$ random data splits for each fusion method, with whiskers showing the minimum and maximum values. In (a), the NLPD of the het-RFF-GP is denoted as a line, as it does not change with $K$.}
\end{figure*}

\subsection{Performance vs number of experts}\label{sec:NLPD_vs_K}
In the first experiment, we fix the number of spectral frequencies to $M = 30$ and perform inference using each method with varying the number of experts $K$. Recall that, while in in BHS and P-BHS, the experts are pre-trained in a separate initial dataset, in MoGPE and PoGPE the experts are jointly inferred with the weights. For the former pair, we further divide the training dataset into two parts, using the first half to train the $K$ experts, and the second half to train the stacking models.
To train the experts, each expert is randomly assigned $\lfloor \frac{N_\text{tr}}{K}\rfloor$ data points, with hyperparameters being selected by type-II maximum likelihood. Note that this means experts receive fewer data points as $K$ increases. The results are shown in Figure \ref{fig_NLPD_vs_experts}.

Our results show that using more experts allows for higher flexibility in MoGPE and PoGPE, with performance increasing as $K$ ranges from $2$ to $5$. MoGPE and PoGPE perform comparably with $K=2$, with MoGPE gaining advantages as more experts are available.
For the stacking methods, the predictive power decreases as the number of experts increases. This can be seen as an artifact of how we trained experts, with a tradeoff between the number of experts and their quality.
The results show that P-BHS outperforms BHS for all tested values of $K$, suggesting that
stacking with log-linear pooling may result in better performance. This is likely due to the property that log-linear pooling with unconstrained weights can arbitrarily adjust the predictive variances.
Intuitively, joint learning (with linear or log-linear pooling) beats stacking in either case.
% On the contrary, the expert models are known for being more successful at integrating weaker learners (cita aqui?). 

\subsection{Performance vs number of spectral frequencies}\label{sec:NLPD_vs_M}

Since the quality of the RFF-GP approximation depends on the number of spectral frequencies $M$, we compare the performance of each method for different values of $M$, with a fixed number of experts $K = 2$. 
The results are shown in Figure~\ref{fig_NLPD_vs_frequencies}. For MoGPE and PoGPE, as well as the het-RFF-GP, there is a strong benefit to increasing the number of spectral frequencies, with the predictive pdf improving by nearly an order of magnitude as $M$ ranges from $10$ to $50$. On the other hand, stacking methods do not benefit much from increasing the number of spectral frequencies; we expect this to be because both experts are already quite accurate, making the stacking GPs rather simple.

\section{Conclusions}
The fusion of GP predictions has received significant attention in the literature, but primarily from the perspective of linear pooling. In the case of log-linear pooling, there has been a lack of a principled method for determining input-dependent weights $\mathbf{w}(\x)$. In this paper, we cast existing approaches under an inclusive framework and introduce novel approaches based on log-linear pooling. We show how the input-dependent weights can be derived in a principled manner in a Bayesian hierarchical model. Furthermore, we analyze the empirical performance of each method on a synthetic example. Future directions of research include additional empirical validation and mathematical analysis of log-linear fusion with unconstrained weights.

% We have reviewed existing approaches to fuse GP predictions. 
% Generally, a fusion model consists of a series of predictive densities of the individual models, and a set of weights for combining them.
% We distinguish two groups, depending on whether these two tasks are carried out separately or jointly. 
% In the first group, a two-stage approach is considered where the GP models/experts are trained first, and then the weights are learned using a second dataset. In the second group, experts and weights are learned together using one dataset. Furthermore, we consider two possible weighted combinations of GP predictions, namely, linear or log-linear pooling. 
% We have compared the performances of the different strategies in several benchmark regression datasets and discussed the strengths and weaknesses of each approach... {\color{red}bla bla bla ...}

\bibliographystyle{IEEEbib.bst}
\bibliography{ref.bib}

\newpage

\end{document}